\documentclass[10pt]{article}
\usepackage[accepted]{rlj} 

\usepackage{amsthm}
\usepackage{newpxtext,newpxmath}

\usepackage{amsmath,amsfonts,bm}

\def\eqref#1{equation~\ref{#1}}

\def\1{\bm{1}}

\def\rq{{\textnormal{q}}}

\def\rva{{\mathbf{a}}}

\def\rvs{{\mathbf{s}}}

\def\rvz{{\mathbf{z}}}

\DeclareMathAlphabet{\mathsfit}{\encodingdefault}{\sfdefault}{m}{sl}
\SetMathAlphabet{\mathsfit}{bold}{\encodingdefault}{\sfdefault}{bx}{n}

\newcommand{\E}{\mathbb{E}}

\newcommand{\ce}{\mathrm{CE}}

\usepackage[ruled,noend]{algorithm2e}
\usepackage{url}
\usepackage{graphicx}
\usepackage{booktabs}
\usepackage[leftbars,xcolor]{changebar}
\theoremstyle{definition}
\newtheorem*{definitionunnumbered}{Definition}

\definecolor{tdcomment}{RGB}{105,98,165}

\SetCommentSty{TDCommentStyle}
\SetKwComment{AlgoCommentLine}{\#\ }{}
\SetKwComment{AlgoRightComment}{$\triangleleft$\ }{}
\newcommand{\CommentLine}[1]{\AlgoCommentLine{#1}}
\newcommand{\RightComment}[1]{\AlgoRightComment*[r]{#1}}
\newcommand{\InputRows}[1]{\begin{tabular}[t]{@{}l@{}}#1\end{tabular}}
\DontPrintSemicolon
\SetArgSty{textnormal}

\title{AMBI: Anytime Model-Based \texorpdfstring{$\pi$}{Policy}-Improvement}
\author{Richard Gao\textsuperscript{1}, Shane Parr\textsuperscript{1}, George Konidaris\textsuperscript{1}}
\emails{}
\affiliations{
$^{1}$\textbf{Brown University}
}
\keywords{model-based reinforcement learning, decision-time planning, anytime algorithms, continuous control}

\begin{document}
\pagestyle{plain}
\maketitle

\begin{abstract}
Model-based reinforcement learning (RL) enables additional computation at decision time to improve policies before executing them. Traditionally, this policy improvement is facilitated by an online planning algorithm that interacts with the model. Using the learned policy as a prior, the planning algorithm specializes it to the decision at the current state. In continuous control, this pattern has demonstrated the potential for improved sample efficiency and better performance that scales with compute. At the same time, this model that enables online planning can also generate simulated experience for updating the agent’s policy or value function via model-free RL. To this end, we introduce Anytime Model-Based $\pi$-Improvement (AMBI), a model-based RL algorithm for continuous control that uses model-free RL for policy improvement at decision time. Specifically, AMBI treats each decision as a reinforcement learning problem rooted at the current state. At each decision, AMBI initializes a local actor and critic from policy and value function priors, updates both on imagined experience from the current state using actor–critic updates, and selects an action with the refined actor. Using actor--critic refinement in place of trajectory optimization, AMBI is competitive with TD-MPC2 on several high-dimensional continuous-control tasks. We also show that AMBI's performance increases smoothly as a function of compute spent during decision time.
\end{abstract}

\section{Introduction}

Among RL algorithms, researchers have long argued that model-based methods hold the key to sample-efficiency in solving complex decision-making problems \citep{sun2019modelbased,chua2018pets,hafner2019planet}. Learning a dynamics model of the environment enables an agent to reason about its decisions without requiring interaction with the world. This is in contrast to model-free RL, which learns exclusively through trial and error in the environment.

Although there is no established consensus on how algorithms should leverage models, two paradigms are prevalent: background planning and decision-time planning \citep{sutton2018reinforcement}. Both methods learn a model and use it to simulate outcomes, but differ in where and how they utilize this imagined experience. Background planning improves a policy or value function that is reused across decisions. In decision-time planning, the agent uses the model after observing the current state to improve its policy for the current decision. This concentrates the model on possible future from that state, at the cost of additional computation before acting.

Historically, model-based methods struggled to match the final returns of strong model-free baselines on continuous-control benchmarks \citep{wang2019benchmarking}. However, decision-time planning with learned models has since yielded strong results in continuous control, from sample-efficient visual control with PlaNet \citep{hafner2019planet} to high-dimensional locomotion and manipulation with TD-MPC2 \citep{hansen2024tdmpc2}. 

In such algorithms. planning algorithms serve as decision-time policy improvement operators. Tree search refines action preferences at the current state \citep{schrittwieser2020muzero}, and methods such as LOOP and TD-MPC2 optimize short action sequences in continuous control \citep{sikchi2022loop,hansen2024tdmpc2}. On the other hand, background planning shows that an agent can update its policy with model-free RL on model-generated experience \citep{sutton1991dyna}. While this is usually done in an offline fashion, it begs the question: can we also use model-free RL to replace the role of planning algorithms as the policy improvement operator at decision time?

To answer this question, we propose \textbf{Anytime Model-Based \(\bm{\pi}\)-Improvement (AMBI)}, a model-based RL algorithm for continuous control that uses model-free RL for policy improvement at decision time. We view each decision as a Markov decision process rooted at the current state, which can be solved by RL. Concretely, AMBI learns a world model, policy, and value function. The policy and value function serve as priors to initialize an off-policy actor--critic learning algorithm that specializes the policy and value function to the current state. To perform policy improvement, AMBI alternates between collecting model rollouts from the current state with the actor and performing policy gradient and TD learning. After sufficient computation, this produces an improved policy for the current state, from which AMBI samples an action to execute, repeating this procedure at the next step.

Prior work has also used RL updates for decision-time planning. RL Search \citep{fickinger2021rlfinetuning} uses true simulator rollouts to fine-tune \emph{pretrained} networks at decision time in discrete games. In Atari, it adapts the policy and value function with on-policy proximal policy optimization \citep{schulman2017proximal}, and in Hanabi, its Q-learning variant adapts a pretrained Q-network. In TreePI's \citep{springenberg2020treepi} brief policy-gradient ablation, they learn their model, but use simulated experience to adapt only the actor under a KL-regularized objective.

Unlike these previous algorithms, AMBI uses imagined experience to specialize both a local actor and critic with off-policy updates before each action, allowing the critic to evaluate the policy as it is updated to drive further improvement. AMBI integrates this procedure into end-to-end training from scratch, using the real experience to train the dynamics model, policy prior, and value prior. Our contributions are as follows:
\begin{enumerate}
    \item We formulate decision-time planning as \emph{Anytime Policy Improvement}, in which an agent uses its available computation to improve a policy from the current state.
    \item We introduce AMBI, an end-to-end model-based RL algorithm for continuous control. AMBI learns its world model, policy prior, and value prior from real interaction and refines a local actor and critic on imagined experience before each action.
    \item We benchmark AMBI on DeepMind Control and Meta-World \citep{tassa2018deepmind,yu2020metaworld}. AMBI is competitive with TD-MPC2 on several tasks using actor--critic refinement in place of trajectory optimization.
    \item On a fixed Humanoid Walk checkpoint, we vary the refinement iterations, actor and critic updates, and rollout horizon. Increasing the refinement budget improves return, and the gains depend strongly on updating the to-go critic.
\end{enumerate}

\section{Background}
\paragraph{Reinforcement Learning}
We model the environment as a Markov decision process (MDP) $\mathcal{M}=\langle \mathcal{S}, \mathcal{A}, \mathcal{P}, \mathcal{R}, \rho_0, \gamma \rangle$, where $\mathcal{S}\subseteq\mathbb{R}^n$ and $\mathcal{A}\subseteq\mathbb{R}^m$ are continuous state and action spaces, $\mathcal{P}:\mathcal{S}\times\mathcal{A}\to\Delta(\mathcal{S})$ is the transition kernel \footnote{$\Delta(\mathcal{X})$ denotes the set of probability measures over $\mathcal{X}$.}, $\mathcal{R}:\mathcal{S}\times\mathcal{A}\to\mathbb{R}$ is the reward function, $\rho_0\in\Delta(\mathcal{S})$ is the initial-state distribution, and $\gamma\in[0,1)$ is the discount factor. The goal of RL is to learn a policy mapping $\pi:\mathcal{S}\to\Delta(\mathcal{A})$ such that $\pi$ maximizes expected discounted return: $\mathcal{J}(\pi) = \E_{\mathcal{M},\pi}\left[{\sum^\infty_{t=0}\gamma^tr_t}\right]$

\paragraph{Model-Based Reinforcement Learning}
In model-based methods, the agent learns a model $\hat{\mathcal{M}}=\langle\hat{\mathcal{P}},\hat{\mathcal{R}}\rangle$ that can generate simulated experience, estimate the consequences of candidate actions. Two common uses of this computation are background planning and decision-time planning \citep{sutton2018reinforcement,alver2024look}.

\emph{Background planning.} The agent uses the model to update a policy or value function that the agent reuses across decisions. Dyna-style methods apply RL updates to simulated transitions, often alongside real transitions. The policy or value function accumulates the results of these updates, so the agent can draw on earlier planning simply by querying the learned function when it acts. The cost of planning is therefore spread across many decisions \citep{sutton2018reinforcement,hamrick2021role}.

\emph{Decision-time planning.} The agent uses the model to help choose an action $\rva_t$ at the current state $\rvs_t$. This is often accomplished through look-ahead search to obtain local value estimates of future states. The agent can use its computation to deepen insights relevant to the observed state, subject to some budget, before acting \citep{alver2024look,fickinger2021rlfinetuning}.


\paragraph{Actor-Critic Algorithms}
Actor--critic methods maintain a parameterized policy (actor) $\pi_\theta$ and value function (critic) $Q_\psi$ \citep{konda1999actorcritic}. Learning interleaves approximate policy evaluation, where the critic follows gradients of a temporal-difference (TD) loss, with approximate policy improvement, where the actor uses gradients to increase the value of its actions according to the critic. For this paper, we specifically care about \emph{off-policy} actor--critic, which can replay experience collected from past policies. \citep{degris2012offpolicy}. Let $\mathcal{B}$ be a replay buffer of transitions $(\rvs,\rva,r,\rvs')$; on collected transitions actor--critic algorithms repeat over the following steps:
\begin{enumerate}
\item \emph{Policy evaluation.} The critic is fitted to a one-step TD target by minimizing
\begin{equation}
\mathcal{L}_{Q}(\psi)
= \E_{(\rvs,\rva,r,\rvs')\sim\mathcal{B}}
\left[D\left(Q_\psi(\rvs,\rva), r + \gamma Q_\psi(\rvs',\rva')\right)\right],
\label{eq:actor-critic-critic-objective}
\end{equation}
where $D$ is a scalar or distributional loss and $\rva'\sim\pi_\theta(\cdot\mid\rvs')$. The TD target is held fixed during the critic update.

\item \emph{Policy improvement.} The actor minimizes the critic-based surrogate loss, while the critic parameters are held constant:
\begin{equation}    
\mathcal{L}_{\pi}(\theta) = -\E_{\rvs\sim\mathcal{B}, \rva\sim\pi_\theta(\cdot\mid\rvs)}\left[Q_\psi(\rvs,\rva)\right].    
\label{eq:actor-critic-actor-objective}
\end{equation}
\end{enumerate}

\section{Anytime Policy Improvement}
\label{sec:anytime-policy-improvement}

Suppose that at the current time $t$, an agent has policy $\pi$, receives observation $\rvs$, and has access to a model $\hat{\mathcal{M}}$. Since the rewards it has already received are fixed, we can view the remaining problem as the original MDP starting at \(s\). In this section, we view policy improvement in terms of solving this sub-MDP, look at how different planning algorithms solve this problem, and justify the use of model-free RL to achieve the same end.

\subsection{The Decision-Time Problem}

 At time $t$, the agent can either act immediately under its current policy $\pi$, or use its model $\hat{\mathcal{M}}$ to improve $\pi$ before acting; the goal of improvement is to maximize expected return from $\rvs$ specifically. We can reframe this optimization problem precisely as maximizing expected return in the sub-MDP rooted at \(\mathbf s\). We call this the to-go MDP \(\mathcal M_{\mathbf s}\).

\begin{changebar}[0.45pt]
\cbcolor{black!55}
\begin{definitionunnumbered}[To-go MDP]
The \emph{to-go MDP} $\mathcal{M}_{\rvs}=\langle \mathcal{S}, \mathcal{A}, \mathcal{P}, \mathcal{R}, \delta_{\rvs}, \gamma \rangle$ is the original MDP $\mathcal{M}$ with its initial-state distribution replaced by the point mass $\delta_{\rvs}$. The expected return of a policy $\pi$ in $\mathcal{M}_{\rvs}$, its \emph{return-to-go}, equals its value at $\rvs$:
\begin{equation}
\label{eq:to-go-objective}
    \mathcal{J}_{\rvs}(\pi)
    \doteq
    \E_{\mathcal{M}_{\rvs},\pi}
    \left[
        \sum_{k=0}^{\infty}\gamma^kr_k
    \right]
    =
    V^\pi_{\mathcal{M}}(\rvs).
\end{equation}
\end{definitionunnumbered}
\end{changebar}

Maximizing the to-go objective is equivalent to optimizing the part of the original objective that remains under the agent's control. The optimal policy for $\mathcal{M}_{\rvs}$ provides the optimal decision rule at $\rvs$ under the original MDP. Thus, we define policy improvement as specializing $\pi$ into a \emph{to-go policy} $\tilde{\pi}$ with higher return-to-go, $\mathcal{J}_{\rvs}(\tilde{\pi}) \geq \mathcal{J}_{\rvs}(\pi)$. 

\subsection{Achieving Policy Improvement}

The learned dynamics and rewards in $\hat{\mathcal{M}}$ let us approximate
this objective by evaluating return-to-go under the model. Let $f$ be any algorithm that specifies how we use the model to improve $\pi$, such as planning algorithms. We use $f$ to seek improvement within the computation available before acting. Decision-time model-based algorithms differ in their choice of $f$. We argue that anytime algorithms fit this role.

\begin{changebar}[0.45pt]
\cbcolor{black!55}
\begin{definitionunnumbered}[Anytime algorithm]
An \emph{anytime algorithm} can return a solution for any allotted computation
budget and use additional computation to further refine the solution
\citep{zilberstein1996using}.
\end{definitionunnumbered}
\end{changebar}

An algorithm $f$ with the anytime property can offer a natural interface between the agent's current policy and its model to achieve policy improvement --- access to $\hat{\mathcal{M}}$ supplies simulated transitions and rewards for candidate actions, while the anytime algorithm $f$ guides which actions and futures are considered next. Crucially, this feedback loop concentrates $f$'s computation on futures relevant to the current decision by letting it examine alternatives and refine those that look promising. Ideally, the quality of the resulting decision improves as $f$ spends more computation at the current state. When $f$'s budget exhausts, the agent uses $\tilde{\pi}(\cdot\mid\rvs)$ to select its action and can replan after observing the next state.

\paragraph{Anytime Policy Improvement} Let $\mathcal{I}^f$ denote the \emph{anytime policy improvement operator}
induced by running anytime algorithm $f$ at the current state using the learned model $\hat{\mathcal{M}}$. Applying it to $\pi$ yields the \emph{to-go policy}
\[
    \tilde{\pi} = \mathcal{I}^f\pi.
\]
The to-go policy $\tilde\pi$ improves on $\pi$ when it achieves higher expected return in the to-go MDP, namely when $\mathcal{J}_{\rvs}(\mathcal{I}^f\pi) > \mathcal{J}_{\rvs}(\pi)$. 

\paragraph{Choice of Anytime Algorithm}
The choice of anytime algorithm $f$ determines how computation is allocated. $\mathcal{I}^\textsc{MCTS}$, as used in
MuZero, updates action estimates along visited branches to guide later
simulations \citep{schrittwieser2020muzero}. In continuous control, sampling-based
trajectory optimization is a common approach to decision-time planning with
learned models. PETS, LOOP, and the TD-MPC family, for example, use the
predicted returns of sampled action sequences to guide subsequent sampling
\citep{chua2018pets,sikchi2022loop,hansen2022tdmpc,hansen2024tdmpc2}.

The to-go MDP suggests that we can use any algorithm designed to maximize return in MDPs for policy improvement --- one such algorithm is RL. Namely, we can let a model-free RL algorithm refine a state-conditioned policy from imagined experience. The benefit is that a shared actor can generalize across the states encountered during lookahead. When different rollouts lead to states that call for similar responses, experience from one rollout can inform how the actor behaves along the others \citep{fickinger2021rlfinetuning}. Furthermore, the policy can utilize first-order gradient information from the return for improvement.

Given that we have policy and value priors, actor--critic algorithms lend us a way to refine the policy incrementally in
continuous action spaces. The critic's gradient with respect to the action
indicates how to adjust a proposed action to increase its estimated value.
A policy-gradient step uses that signal to refine the actor, and critic
updates let the action values reflect the actor's changing behavior.
This makes actor--critic a natural and appealing anytime choice. The agent starts with a learned actor it can use immediately and can spend additional computation refining it on imagined experience, including transitions from earlier rollouts, acting with the latest policy when the budget runs out. This style of updates is reminiscent of background planning methods, where model-free updates are applied to imagined experience. However, unlike background planning, we focus this computation on the current state specifically.


\section{Anytime Model-Based \texorpdfstring{$\pi$}{Policy}-Improvement}
\label{sec:ambi}

In this section, we propose \textbf{Anytime Model-Based \texorpdfstring{$\bm{\pi}$}{Policy}-Improvement (AMBI)}, a model-based RL algorithm for continuous control that uses actor--critic as Anytime Policy Improvement. Across decisions, AMBI maintains a world model, policy prior, and value prior. At decision time, the policy and value priors are specialized to the current state by instantiating a to-go actor and critic that are subsequently updated using imagined transitions generated from the current state. After sufficient updates, AMBI takes a step in the real environment with the final to-go actor. Real transitions then train the world model, policy prior, and value prior, reused at later decisions. 

Before diving into policy improvement, we briefly overview the world model, value prior, and policy prior: AMBI learns a world model $\hat{\mathcal{M}}=\langle \mathcal{E}_\phi, \mathcal{P}_\phi, \mathcal{R}_\phi\rangle$, a policy prior $\pi_\theta$, and a value prior $Q_\psi$ from real replay experience. $\hat{\mathcal{M}}$ is a task-oriented latent dynamics (TOLD) model \citep{hansen2022tdmpc}, comprising the encoder $\mathcal{E}_\phi$, latent dynamics predictor $\mathcal{P}_\phi$, and reward predictor $\mathcal{R}_\phi$.
\begin{equation}
\label{eq:ambi-components}
\begin{array}{@{}l@{\qquad}l@{\;}l@{}}
    \text{Encoder}          & \rvz_i = \mathcal{E}_\phi(\rvs_i) \\
    \text{Latent predictor} & \rvz_{i+1} = \mathcal{P}_\phi(\rvz_i, \rva_i)
    & \smash{\left.\vphantom{\begin{array}{@{}c@{}}
        \mathstrut \\ \mathstrut \\ \mathstrut
      \end{array}}\right\}}\;\text{World model} \\
    \text{Reward predictor} & \hat{r}_i = \mathcal{R}_\phi(\rvz_i, \rva_i) \\[0.4em]
    \text{Policy prior}     & \hat{\rva}_i \sim \pi_\theta(\cdot \mid \rvz_i) \\
    \text{Value prior}      & \hat{q}_i = Q_\psi(\rvz_i, \rva_i)
\end{array}
\end{equation}
To start a rollout, AMBI encodes $\rvs_t$ as $\rvz_t=\mathcal{E}_\phi(\rvs_t)$. For each action sampled from the to-go actor, the world model predicts the reward and next latent state. The policy and value priors provide the initial parameters for the to-go actor and critic.

In Section~\ref{sec:sac-based-api}, we illustrate the decision-time refinement procedure, then in Section~\ref{sec:model-and-control-prior-learning}, we explain how AMBI learns its world model, policy prior, and value prior from real interaction.

\subsection{\texorpdfstring{$\mathcal{I}^{\text{\textsc{AC}}}$: }{}Actor--Critic as Anytime Policy Improvement}
\label{sec:sac-based-api}

At time $t$, AMBI explicitly represents its to-go policy with a parameterized actor: $\tilde{\pi}=\pi_{\tilde{\theta}}$. We instantiate a to-go  actor $\pi_{\tilde{\theta}}$, critic $Q_{\tilde{\psi}}$, and empty replay $\tilde{\mathcal{B}}$; $\tilde{\theta}$ and $\tilde{\psi}$, are initialized from policy and value prior parameters $\theta$ and $\psi$, respectively.

\paragraph{$H$-Step Objective}
We seek a to-go policy $\pi_{\tilde{\theta}}$ that maximizes the return-to-go
$\mathcal{J}_{\rvs_t}(\pi_{\tilde{\theta}})$ (Equation~\ref{eq:to-go-objective}).
However, estimating the full return by simulation is costly at decision time, and model errors can compound over
long rollouts \citep{hansen2022tdmpc}. We therefore approximate the to-go objective
$\mathcal{J}_{\rvs_t}(\pi_{\tilde{\theta}})$ with $H$-step improvement. Beyond $H$, we bootstrap the remaining return using our to-go critic. The resulting optimization is a search over to-go policy parameters $\tilde\theta$,
\begin{equation}
\label{eq:h-step-to-go-objective}
\max_{\tilde{\theta}}\quad
\E_{\hat{\mathcal{M}},\pi_{\tilde{\theta}}}\!\left[
    \gamma^H Q_{\tilde{\psi}}(\rvz_{t+H},\rva_{t+H}) +
    \sum_{k=0}^{H-1}\gamma^k\mathcal{R}_\phi(\rvz_{t+k},\rva_{t+k})
\right]
\end{equation}
under rolled-out latent dynamics. The to-go critic $Q_{\tilde{\psi}}$ evaluates the final state--action pair.
Shorter rollouts place more weight on the value estimate and rely less on long-horizon model predictions \citep{sikchi2022loop}.

\paragraph{Actor--Critic Updates}
AMBI collects $H$-step rollouts from $\rvz_t$ under the current to-go actor
and stores their transitions in $\tilde{\mathcal{B}}$. To improve the to-go policy for the objective in Equation~\ref{eq:h-step-to-go-objective}, AMBI draws minibatches from replay and updates the critic before the actor. The updated actor generates
the next batch of rollouts, while replay lets both updates reuse previously imagined transitions.

\emph{Policy evaluation.} We fit the to-go critic to a one-step TD target for the current actor by minimizing
\begin{equation}
\label{eq:sac-api-critic-objective}
\mathcal{L}_{Q}(\tilde{\psi})
= \E_{(\rvz,\rva,\hat{r},\rvz')\sim\tilde{\mathcal{B}}}
\left[\ce\left(
    Q_{\tilde{\psi}}(\rvz,\rva),
    \hat{r}+\gamma Q_{\tilde{\psi}^{-}}(\rvz',\rva')
\right)\right],
\end{equation}
where $\ce$ is the cross-entropy prediction loss and
$\rva'\sim\pi_{\tilde{\theta}}(\cdot\mid\rvz')$. Sampling the next action
from the latest actor lets the critic track the to-go policy as it changes.
At each decision, we initialize the target critic as a copy of the to-go
critic and update it by exponential averaging.

\emph{Policy improvement.} We update the to-go actor by minimizing
\begin{equation}
\label{eq:sac-api-actor-objective}
\mathcal{L}(\tilde{\theta})
= -\E_{\rvz\sim\tilde{\mathcal{B}},\,\rva\sim\pi_{\tilde{\theta}}(\cdot\mid\rvz)}\!\left[
    Q_{\tilde{\psi}}(\rvz,\rva) /\tilde{S}
    + \alpha \mathcal{H}\!\left(\pi_{\tilde{\theta}}(\cdot\mid\rvz)\right)
\right],
\end{equation}
where $\mathcal{H}$ denotes the scaled policy entropy and $\alpha$ is its
fixed weight. The $Q$ term favors actions with high estimated return, and
the entropy bonus encourages the actor to sample alternatives in subsequent
rollouts. The local scale $\tilde{S}$ tracks the percentile range of $Q$
values used in the actor loss \citep{hafner2023dreamerv3}.
At each decision, AMBI initializes $\tilde{S}\gets S$ from the running
scale maintained during prior learning. Before each actor update, let
$\mathbf{q}$ contain the values $Q_{\tilde{\psi}}(\rvz,\rva)$ for the
state--action samples in Equation~\ref{eq:sac-api-actor-objective}. We update
the local scale as
\begin{equation}
\label{eq:decision-time-scale-ema}
\tilde{S} \doteq \operatorname{EMA}\!\Bigl(
    \max\!\bigl\{1,\,\operatorname{Per}(\mathbf{q},95)
    - \operatorname{Per}(\mathbf{q},5)\bigr\},\,1-\tau_S
\Bigr),
\end{equation}
where $\operatorname{Per}(\mathbf{q},p)$ is the $p$-th percentile of
$\mathbf{q}$ and $1-\tau_S$ is the EMA decay.

\paragraph{Exploration and Exploitation}
The balance between exploration and exploitation in the actor's loss (Equation~\ref{eq:sac-api-actor-objective}) is driven by how $\tilde{S}$ moves during optimization. The outer scale $S$ tracks $Q$ values for states sampled from real replay
$\mathcal{B}$ and actions drawn from the policy prior $\pi_\theta$.
These states span many decisions and have wider expected
returns. States sampled from the to-go replay $\tilde{\mathcal{B}}$ come
from short rollouts rooted at $\rvz_t$, so their $Q$ values can span a
narrower range. When the to-go minibatches yield a smaller scale than $S$, the inherited
scale gives the entropy bonus more influence in the early actor updates.
As $\tilde{S}$ falls toward the local range, the updates put more weight
on the critic's estimates, shifting the balance toward exploitation.

AMBI repeats this cycle for $J$ iterations, generating each new batch of
rollouts from $\rvz_t$ with the latest actor.
Training on imagined transitions rooted at the same state progressively
specializes the actor and critic to the current decision. After the final
iteration, AMBI uses the to-go actor $\pi_{\tilde\theta}$ at $\rvz_t$ to select $\rva_t$. Algorithm~\ref{alg:api} summarizes the refinement procedure $\mathcal{I}^{\textsc{AC}}$.

\begin{algorithm}[tbp]
\caption{$\mathcal{I}^{\textsc{AC}}$: Anytime Policy Improvement with Actor--Critic}
\label{alg:api}
\KwIn{\InputRows{%
       current state: $\rvs_t$\tabularnewline
       world model: $\mathcal{E}_\phi,\mathcal{P}_\phi,\mathcal{R}_\phi$\tabularnewline
       policy and value priors: $\pi_\theta,Q_\psi$\tabularnewline
       budget: iterations $J$, rollouts $R$, max depth $H$, actor updates $N_{\pi}$, critic updates $N_{Q}$}}
\KwOut{Improved to-go policy $\pi_{\tilde{\theta}}$}

$\tilde{\theta},\tilde{\psi} \gets \theta,\psi$\RightComment{Initialize to-go actor and critic from priors}
$\tilde{\mathcal{B}} \gets \emptyset$\RightComment{Initialize to-go replay}
$\rvz_t \gets \mathcal{E}_\phi(\rvs_t)$\RightComment{Encode root state}

\CommentLine{Generate imagined experience and improve the to-go policy}
\For{iteration $j=1,\ldots,J$}{
    \For{rollout $r=1,\ldots,R$}{
        \For{rollout step $\tau=t,\ldots,t+H-1$}{
            $\rva_\tau \sim \pi_{\tilde{\theta}}(\cdot\mid\rvz_\tau)$ \\
            $\rvz_{\tau+1} \gets \mathcal{P}_\phi(\rvz_\tau,\rva_\tau)$ \\
            $\hat{r}_\tau \gets \mathcal{R}_\phi(\rvz_\tau,\rva_\tau)$ \\
            $\tilde{\mathcal{B}} \gets \tilde{\mathcal{B}} \cup
            \{(\rvz_\tau,\rva_\tau,\hat{r}_\tau,\rvz_{\tau+1})\}$ \\
        }
    }
    \CommentLine{Update the to-go actor and critic from replay}
    \For{critic update $n=1,\ldots,N_{Q}$}{
        $(\rvz,\rva,\hat{r},\rvz')\sim\tilde{\mathcal{B}}$ \\
        $\tilde{\psi} \gets \tilde{\psi}
        -\eta_{\tilde{\psi}}\nabla_{\tilde{\psi}}\mathcal{L}_{Q}(\tilde{\psi})$\RightComment{Critic loss, Equation~\ref{eq:sac-api-critic-objective}}
    }
    \For{actor update $n=1,\ldots,N_{\pi}$}{
        $\rvz\sim\tilde{\mathcal{B}}$ \\
        $\tilde{\theta} \gets \tilde{\theta}
        -\eta_{\tilde{\theta}}\nabla_{\tilde{\theta}}\mathcal{L}(\tilde{\theta})$\RightComment{Actor loss, Equation~\ref{eq:sac-api-actor-objective}}
    }
}

\Return{$\pi_{\tilde{\theta}}$}\RightComment{Sample improved root action}
\end{algorithm}

\subsection{Model and Prior Learning} \label{sec:model-and-control-prior-learning}

At state $\rvs_t$, anytime policy improvement refines the policy prior into
the to-go policy $\pi_{\tilde{\theta}}=\mathcal{I}^{\text{\textsc{AC}}}\pi_\theta$ with actor--critic.
During training, AMBI samples
$\rva_t\sim\pi_{\tilde{\theta}}(\cdot\mid\rvz_t)$ and executes it in the
environment. The resulting transition $(\rvs_t,\rva_t,r_t,\rvs_{t+1})$ is
stored in the real replay buffer $\mathcal{B}$. We sample short sequences
from this buffer to train the world model, policy prior, and value prior
using the same learning procedure as TD-MPC2 \citep{hansen2024tdmpc2}.

\paragraph{World model and value prior}
For a sequence $\tau=(\rvs_i,\rva_i,r_i,\rvs_{i+1})_{i=0}^{K-1}$ sampled
from $\mathcal{B}$, we encode the first state as $\rvz_0=\mathcal{E}_\phi(\rvs_0)$
and roll the latent dynamics forward under the recorded actions,
$\rvz_{i+1}=\mathcal{P}_\phi(\rvz_i,\rva_i)$. This trains the model to make
predictions from its own earlier predictions, as it does during decision-time
rollouts. We jointly train the world model and value prior with a weighted sum
of latent consistency, reward prediction, and TD losses,
\begin{equation}
\label{eq:persistent-model-value-objective}
\mathcal{L}(\phi,\psi)
= \frac{1}{K}\,
  \E_{\tau\sim\mathcal{B}}\!\left[
    \sum_{i=0}^{K-1}\lambda^i\Bigl(
      c_z\,\bigl\|\rvz_{i+1}-[\bar{\rvz}_{i+1}]_{\mathrm{sg}}\bigr\|_2^2
      +c_r\,\ce(\hat{r}_i,r_i)
      +c_q\,\ce\bigl(\hat{q}_i,[q_i]_{\mathrm{sg}}\bigr)
    \Bigr)
  \right],
\end{equation}
where $\hat{r}_i$ and $\hat{q}_i$ are the predictions in Equation~\ref{eq:ambi-components}.
The consistency term compares the predicted latent with the encoding of the
observed next state, $\bar{\rvz}_{i+1}=\mathcal{E}_\phi(\rvs_{i+1})$. The TD
target is $q_i=r_i+\gamma Q_{\psi^-}(\bar{\rvz}_{i+1},\rva'_{i+1})$, with
$\rva'_{i+1}\sim\pi_\theta(\cdot\mid\bar{\rvz}_{i+1})$. Here $[\cdot]_{\mathrm{sg}}$ denotes
stop-gradient, and $\ce$ denotes the cross-entropy prediction loss.
The coefficients $c_z,c_r,c_q$ weight the three terms,
and $\lambda\in(0,1]$ controls the weighting across rollout depths.

\paragraph{Policy prior}
We update $\pi_\theta$ separately on the unrolled latent states, including the
final latent $\rvz_K$. At each latent state, we sample
$\hat{\rva}_i\sim\pi_\theta(\cdot\mid\rvz_i)$ and use the same Q-normalized,
entropy-regularized objective as the to-go actor in
Equation~\ref{eq:sac-api-actor-objective},
\begin{equation}
\label{eq:persistent-policy-objective}
\mathcal{L}_\pi(\theta)
= -\E_{\tau\sim\mathcal{B}}\!\left[
    \frac{1}{K+1}\sum_{i=0}^{K}\lambda^i\left(
        \frac{Q_\psi(\rvz_i,\hat{\rva}_i)}{S}
        +\alpha\mathcal{H}\!\left(\pi_\theta(\cdot\mid\rvz_i)\right)
    \right)
\right].
\end{equation}
This update holds the latent states and value prior fixed. The normalization
uses the running scale $S$ maintained during prior learning. We update $S$
before each policy-prior step using the percentile-range EMA in
Equation~\ref{eq:decision-time-scale-ema}, with $\mathbf{q}$ containing
$Q_\psi(\rvz_0,\hat{\rva}_0)$ for the first latent state of each sampled
replay sequence. At the next
decision, $\pi_\theta$ and $Q_\psi$ initialize the to-go actor and critic,
and $S$ initializes $\tilde{S}$.

\paragraph{Architecture and implementation}
We follow TD-MPC2's architecture for the world model and priors. For state
observations, all components in Equation~\ref{eq:ambi-components} are MLPs
with LayerNorm and Mish in their hidden layers. To help mitigate exploding
gradients during training, the encoder and latent dynamics bound their
512-dimensional outputs with SimNorm, applying softmax within each group of
eight coordinates \citep{hansen2024tdmpc2}.
The policy outputs the mean and bounded log standard deviation of a diagonal
Gaussian. We draw a reparameterized sample and apply $\tanh$ to obtain a
bounded action. The reward predictor and each critic head output logits over
101 bins in symlog space and are trained with cross-entropy against two-hot
targets. Symlog compresses large magnitudes of either sign, allowing a fixed
set of bins to cover a wide range of rewards and returns
\citep{hafner2023dreamerv3}. We decode each critic head to a scalar $Q$ before
aggregating the ensemble. Actor updates can favor actions with overestimated
$Q$ values, and TD bootstrapping can propagate those errors
\citep{fujimoto2018td3}. Because AMBI uses actor--critic updates for both
prior learning and decision-time refinement, we use a pessimistic bootstrap
in both cases, taking the minimum over all ten heads \citep{chang2026mrsq}.
Actor updates use the mean over all ten.

\begin{algorithm}[!bp]
\caption{AMBI: Anytime Model-Based \texorpdfstring{$\pi$}{Policy}-Improvement}
\label{alg:ambi}
\KwIn{\InputRows{%
       World model: $(\mathcal{E}_\phi,\mathcal{P}_\phi,\mathcal{R}_\phi)$; $\phi$ randomly initialized\tabularnewline
       Policy and value priors: $(\pi_\theta,Q_\psi)$; $\theta,\psi$ randomly initialized\tabularnewline
       Training settings: episode length $T$, $N_{\text{update}}$ updates, and replay unroll length $K$\tabularnewline
       Optimization: learning rates $\eta_\phi,\eta_\psi,\eta_\theta$; target rate $\tau_Q$}}

$\mathcal{B}\gets\emptyset$\RightComment{Initialize real replay}
$\psi^-\gets\psi,\quad S\gets 1$\RightComment{Initialize target critic and Q scale}

\While{not converged}{
    \CommentLine{Collect an episode using Algorithm~\ref{alg:api}}
    \For{time $t=1,\ldots,T$}{
        $\pi_{\tilde{\theta}} \gets \mathcal{I}^{\text{\textsc{AC}}}\pi_\theta$ \RightComment{Anytime Policy Improvement}
        $\rva_t \sim \pi_{\tilde{\theta}}(\cdot\mid\mathcal{E}_\phi(\rvs_t))$ \RightComment{Sample action from to-go policy}
        $\rvs_{t+1}\sim p(\cdot\mid\rvs_t,\rva_t),\quad
        r_t\gets\mathcal{R}(\rvs_t,\rva_t)$\RightComment{Step environment}
        $\mathcal{B}\gets\mathcal{B}\cup
        \{(\rvs_t,\rva_t,r_t,\rvs_{t+1})\}$\RightComment{Add transition to replay}
    }

    \CommentLine{Update the world model, policy prior, and value prior from replay}
    \For{gradient update $u=1,\ldots,N_{\text{update}}$}{
        $(\rvs_i,\rva_i,r_i,\rvs_{i+1})_{i=0}^{K-1}\sim\mathcal{B}$\RightComment{Sample training sequence}
        $\rvz_0\gets\mathcal{E}_\phi(\rvs_0)$\RightComment{Encode first observation}
        \For{rollout depth $i=0,\ldots,K-1$}{
            $\hat{r}_i\gets\mathcal{R}_\phi(\rvz_i,\rva_i)$ \\
            $\hat{\rq}_i\gets Q_\psi(\rvz_i,\rva_i)$ \\
            $\rvz_{i+1}\gets\mathcal{P}_\phi(\rvz_i,\rva_i)$ \\
            $\hat{\rva}_i\sim\pi_\theta(\cdot\mid\rvz_i)$ \\
        }
        $\hat{\rva}_K\sim\pi_\theta(\cdot\mid\rvz_K)$\;
        \CommentLine{Update from rolled-out estimates (Equations~\ref{eq:persistent-model-value-objective} and~\ref{eq:persistent-policy-objective})}
        $\phi\gets\phi-\eta_\phi\nabla_\phi\mathcal{L}(\phi,\psi)$\RightComment{Update world model}
        $\psi\gets\psi-\eta_\psi\nabla_\psi\mathcal{L}(\phi,\psi)$\RightComment{Update value prior}
        Update $S$ using policy $Q$ values at $\rvz_0$\RightComment{Update Q scale}
        $\theta\gets\theta-\eta_\theta\nabla_\theta\mathcal{L}_\pi(\theta)$\RightComment{Update policy prior}
        $\psi^-\gets(1-\tau_Q)\psi^-+\tau_Q\psi$\RightComment{Update target critic}
    }
}
\end{algorithm}

\section{Experiments}
\label{sec:experiments}

We evaluate AMBI on continuous-control tasks to study how it performs when trained end to end, how much decision-time refinement improves on its learned policy prior, and how that improvement depends on the computation available before each action.

\subsection{Experimental Setup}

\paragraph{Benchmarks.}
We benchmark AMBI on seven challenging high-dimensional continuous-control tasks from the DeepMind Control Suite \citep{tassa2018deepmind,hansen2024tdmpc2}. The Dog and Humanoid locomotion tasks require coordinating many actuators to balance and move, with 38- and 21-dimensional action spaces, respectively. We also evaluate on Assembly, Hand Insert, Push, and Stick Pull from Meta-World \citep{yu2020metaworld}. We train a separate agent from state observations on each task and report undiscounted episode return for locomotion and final-step success rate for manipulation.

\paragraph{Baselines and Evaluation.}
We compare against the single-task TD-MPC2, SAC, and DreamerV3 results released with TD-MPC2 \citep{hansen2024tdmpc2,haarnoja2018softapplications,hafner2023dreamerv3}. We index all curves by training decisions, with each decision corresponding to two simulator steps. AMBI uses $J=8$ refinement iterations per decision. Each iteration generates $R=128$ rollouts of horizon $H=3$ and performs $N_Q=16$ critic updates and $N_\pi=4$ actor updates. At evaluation, AMBI executes the squashed Gaussian mean. Each checkpoint is evaluated over ten episodes per training seed. We evaluate frozen checkpoints every 50,000 decisions on DMControl and use evaluations collected during training every 25,000 decisions on Meta-World. See Appendix~\ref{app:hyperparameters} for the hyperparameters.

\subsection{Benchmark Performance}

\begin{figure}[tbp]
    \centering
    \includegraphics[width=\linewidth]{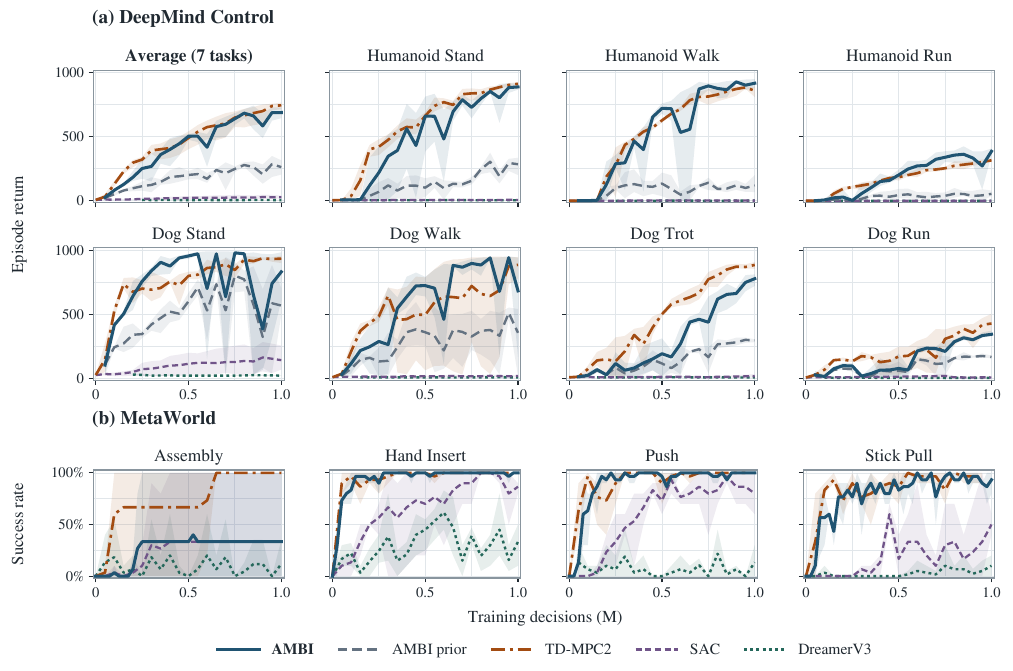}
    \caption{\textbf{Continuous-control benchmarks.} (a) DMControl episode return. (b) Meta-World final-step success. Lines and bands show three-seed means and pointwise 95\% bootstrap intervals.}
    \label{fig:ambi-benchmarks}
\end{figure}

On DMControl, refinement improves on the learned policy prior across the locomotion suite. AMBI exceeds TD-MPC2 on Humanoid Walk and Run at the end of training (Figure~\ref{fig:ambi-benchmarks}a). On Meta-World, AMBI achieves higher average success than SAC and DreamerV3. TD-MPC2 leads overall, with the largest gap on Assembly (Figure~\ref{fig:ambi-benchmarks}b).

\subsection{Decision-Time Budget}

\begin{figure}[tbp]
    \centering
    \includegraphics[width=0.90\linewidth]{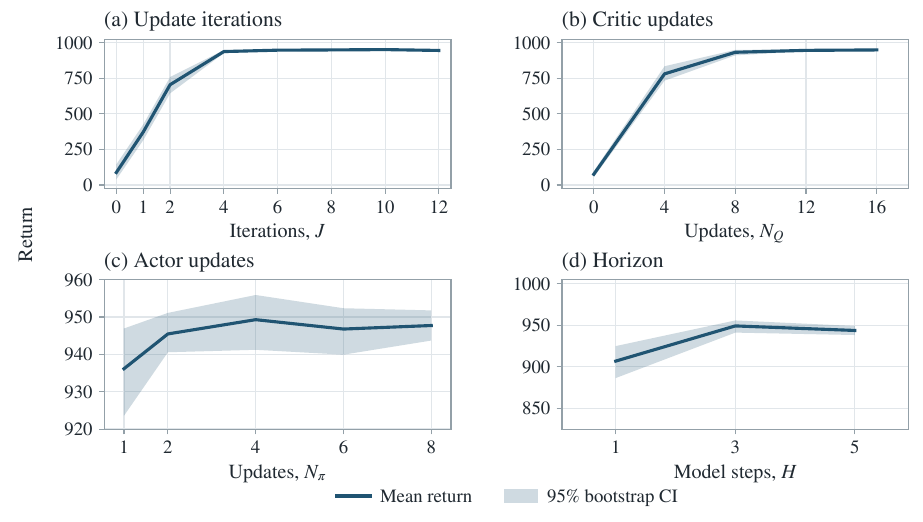}
    \caption{\textbf{Decision-time budget on Humanoid Walk.} Each panel varies one setting at a fixed checkpoint. Lines and shading show means and pointwise 95\% bootstrap confidence intervals over ten matched episodes.}
    \label{fig:ambi-budget}
\end{figure}

We vary one decision-time setting at a time on a fixed Humanoid Walk checkpoint trained for one million decisions \citep{hansen2022tdmpc}. The agent recovers most of the benefit of refinement using fewer iterations than it used during training (Figure~\ref{fig:ambi-budget}a). Return is most sensitive to the number of critic updates. Refining the actor with a fixed critic gives little improvement over the prior, and varying the actor budget or horizon has a smaller effect (Figure~\ref{fig:ambi-budget}b--d).

\begin{figure}[!htbp]
    \centering
    \includegraphics[width=0.59\linewidth]{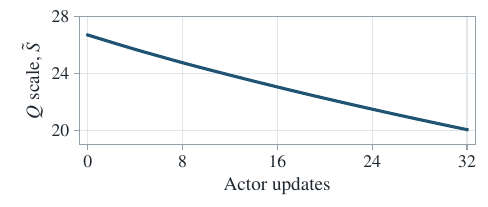}
    \caption{\textbf{Local $Q$ scale} $\tilde{S}$ across actor updates within one Humanoid Walk decision.}
    \label{fig:ambi-inner-scale}
\end{figure}

\paragraph{Exploration and Exploitation.}
In the Humanoid Walk example, the local scale $\tilde{S}$ falls during refinement, increasing the weight on the critic's estimates relative to the entropy bonus (Figure~\ref{fig:ambi-inner-scale}, Equation~\ref{eq:sac-api-actor-objective}).

\section{Related Work}
\label{sec:related-work}

\paragraph{Trajectory optimization.}
PETS and PlaNet use learned models to optimize action sequences \citep{chua2018pets,hafner2019planet}, and LOOP and TD-MPC supplement short-horizon search with a learned terminal value \citep{sikchi2022loop,hansen2022tdmpc}. Dream-MPC differentiates through latent dynamics to optimize the action sequence \citep{spieler2026dreammpc}. AMBI builds on TD-MPC2's latent model \citep{hansen2024tdmpc2} and learns a to-go policy that can respond to the states encountered within an imagined rollout.

\paragraph{Learning from planning.}
Expert iteration trains a policy to imitate decisions produced by search \citep{anthony2017expert}. Planner amortization, BMPC, and BOOM apply this idea to continuous control by training a persistent actor to reproduce or align with MPC behavior \citep{byravan2022amortization,wang2025bmpc,zhan2025boom}. In AMBI, decision-time refinement shapes the real experience collected by the agent, and the persistent priors learn from that experience.

\paragraph{Decision-time policy refinement.}
TD Search and Policy Gradient Search use simulated experience to update a value function or rollout policy during search \citep{silver2012tdsearch,anthony2019pgsearch}. In continuous control, POPLIN-P searches over policy parameters \citep{wang2020poplin}, and TreePI's policy-gradient variant temporarily adapts the actor on learned-model rollouts with a fixed terminal value estimator \citep{springenberg2020treepi}. RL Search's Atari variant adapts both the policy and value network with PPO on emulator rollouts \citep{fickinger2021rlfinetuning}. ITPO differentiates through a learned model to adapt an offline-trained actor \citep{deb2026itpo}, and \citet{majumdar2025dyna} use current-state model rollouts to update a persistent actor and critics with TD3. At each decision, AMBI initializes a fresh actor and critic from learned priors and refines both off-policy using imagined replay rooted at the current state.

\section{Conclusion}
\label{sec:conclusion}

We cast each decision as a to-go MDP and used AMBI to test whether model-free RL could improve a continuous-control policy at decision time. AMBI improved on its learned policy prior across the locomotion suite and was competitive with TD-MPC2 on several tasks. The ablations showed that additional refinement improved return and that critic updates were important to those gains. These results show that actor--critic learning is a viable alternative to trajectory optimization for decision-time policy improvement.

With a finite decision-time budget, AMBI may miss useful behaviors that the policy prior rarely explores. Errors in the learned model or critic can make gains in the imagined objective fail to translate to real return. At decision time, sampling-based planners such as TD-MPC2 use terminal $Q$ values in the scores that rank and weight candidate trajectories \citep{hansen2024tdmpc2}. AMBI's actor updates also depend on the critic's action gradients, which can be misleading even when the critic ranks sampled actions correctly \citep{doro2020mage}. Future work could diversify imagined rollouts and train critics that provide more reliable action gradients.

\bibliography{main}
\bibliographystyle{rlj}

\clearpage
\appendix

\section{Hyperparameters}
\label{app:hyperparameters}

\begin{table}[!htbp]
    \centering
    \caption{\textbf{AMBI hyperparameters.} Settings for the ten-critic configuration. Suite-specific values are listed in DMControl / Meta-World order.}
    \label{tab:ambi-hyperparameters}
    \small
    \renewcommand{\arraystretch}{1.04}
    \begin{tabular}{@{}l@{\hspace{2em}}l@{}}
        \toprule
        \textbf{Hyperparameter} & \textbf{Value} \\
        \midrule
        \multicolumn{2}{@{}l@{}}{\textcolor[HTML]{205472}{\textbf{\underline{Decision-time policy improvement}}}} \\
        Imagined rollout horizon ($H$) & $3$ \\
        Refinement iterations per decision ($J$) & $8$ \\
        Rollouts per iteration ($R$) & $128$ \\
        Critic updates per iteration ($N_Q$) & $16$ \\
        Actor updates per iteration ($N_\pi$) & $4$ \\
        To-go batch size & $256$ \\
        \addlinespace[4pt]
        \multicolumn{2}{@{}l@{}}{\textcolor[HTML]{205472}{\textbf{\underline{Policy priors and to-go actors}}}} \\
        Gaussian log standard deviation bounds & $[-10,2]$ \\
        Entropy coefficient ($\alpha$) & $2.1\times 10^{-3}$ / $4\times 10^{-4}$ \\
        \addlinespace[4pt]
        \multicolumn{2}{@{}l@{}}{\textcolor[HTML]{205472}{\textbf{\underline{Replay buffers}}}} \\
        Real replay capacity & $1{,}000{,}000$ transitions \\
        Real replay sampling & Uniform sequences \\
        To-go replay capacity & $3{,}072$ transitions \\
        To-go replay sampling & Uniform with replacement \\
        \addlinespace[4pt]
        \multicolumn{2}{@{}l@{}}{\textcolor[HTML]{205472}{\textbf{\underline{Architecture}}}} \\
        Encoder hidden layers and width & $1 \times 256$ \\
        Other MLP hidden layers and width & $2 \times 512$ \\
        Latent state dimension & $512$ \\
        Hidden layer normalization and activation & LayerNorm and Mish \\
        SimNorm group size & $8$ \\
        Critic heads & $10$ \\
        Critic dropout rate & $1\%$ \\
        Ensemble reductions (TD target / actor) & Minimum / mean over all heads \\
        Reward and value bins & $101$ over $[-10,10]$ in symlog space \\
        \addlinespace[4pt]
        \multicolumn{2}{@{}l@{}}{\textcolor[HTML]{205472}{\textbf{\underline{Optimization}}}} \\
        Real-replay update-to-data ratio & $1$ \\
        Real-replay batch size & $256$ \\
        Training unroll length ($K$) & $3$ \\
        Latent consistency loss coefficient & $20$ \\
        Reward prediction loss coefficient & $0.1$ \\
        Value prediction loss coefficient & $0.1$ \\
        Temporal loss weighting ($\lambda$) & $0.5$ \\
        Target critic update rate ($\tau_Q$) & $0.01$ \\
        Actor $Q$ normalization & EMA of $5$th--$95$th percentile range, floor $1$ \\
        $Q$-scale update rate ($\tau_S$) & $0.01$ \\
        Optimizer & Adam \\
        Adam $\epsilon$ (actors / other parameters) & $10^{-5}$ / $10^{-8}$ \\
        Learning rate (other parameters) & $3\times 10^{-4}$ \\
        Encoder learning rate & $9\times 10^{-5}$ \\
        Gradient clipping norm & $20$ \\
        Discount factor ($\gamma$) & $0.99$ / $0.95$ \\
        Random warm-up decisions & $2{,}500$ / $1{,}000$ \\
        Initial replay updates & $2{,}500$ / $1{,}000$ \\
        \bottomrule
    \end{tabular}
\end{table}

\end{document}